\documentclass[twoside]{article}

\usepackage[utf8]{inputenc} % allow utf-8 input
\usepackage[T1]{fontenc}    % use 8-bit T1 fonts
\usepackage[colorlinks, citecolor=blue]{hyperref}
\usepackage{booktabs}       % professional-quality tables
\usepackage{nicefrac}       % compact symbols for 1/2, etc.
\usepackage{microtype}      % microtypography
\usepackage{graphicx}
\usepackage{subfigure}
\usepackage{amsfonts, amsmath,amssymb,amsthm}
\usepackage{xcolor}
\usepackage{macros}
\usepackage{stmaryrd}
\usepackage{multirow}
\usepackage{enumitem}
\usepackage{xr}
\newtheorem{proposition}{Proposition}

\newcommand{\myparagraph}[1]{\vspace{-7pt}\paragraph{{#1.}}}
\newcommand\sN{\mathcal{N}}

\usepackage[accepted]{aistats2019}

\begin{document}

\setlength{\belowdisplayskip}{4pt}
\setlength{\abovedisplayskip}{4pt}

\twocolumn[

\runningtitle{Defending against Whitebox Adversarial Attacks}
\aistatstitle{Defending against Whitebox Adversarial Attacks \\ via Randomized Discretization}

\aistatsauthor{ Yuchen Zhang  \And  Percy Liang }

\aistatsaddress{ \tt{yuczhan@microsoft.com} \\Microsoft Corporation\\Berkeley, CA 94704, USA \And \tt{pliang@cs.stanford.edu} \\Computer Science Department\\Stanford University, CA 94305, USA } ]

\begin{abstract}
Adversarial perturbations dramatically decrease the accuracy of state-of-the-art image classifiers. In this paper, we propose and analyze a simple and computationally efficient defense strategy: inject random Gaussian noise, discretize each pixel, and then feed the result into any pre-trained classifier. Theoretically, we show that our randomized discretization strategy reduces the KL divergence between original and adversarial inputs, leading to a lower bound on the classification accuracy of any classifier against any (potentially whitebox) $\ell_\infty$-bounded adversarial attack. Empirically, we evaluate our defense on adversarial examples generated by a strong iterative PGD attack. On ImageNet, our defense is more robust than adversarially-trained networks and the winning defenses of the NIPS 2017 Adversarial Attacks \& Defenses competition.
\end{abstract}

\section{Introduction}

% Intro to adversarial ML
Machine learning models have achieved impressive success in diverse tasks, but
many of them are sensitive to small perturbations of the input. Recent studies
show that adversarially constructed perturbations, even if inperceptibly small,
can dramatically decrease the accuracy of state-of-the-art models in image
classification~\cite{szegedy2013intriguing, goodfellow2014explaining,
kurakin2016world, liu2016delving}, face
recognition~\cite{sharif2016accessorize}, robotics~\cite{melis2017deep}, speech
recognition~\cite{cisse2017houdini} and malware
detection~\cite{biggio2013evasion}. Such vulnerability exposes security
concerns and begs for a more reliable way to build ML systems.

% Training time can work in whitebox, but fail to scale
Many techniques have been proposed to robustify models for image
classification. A mainstream approach is to train the model on adversarial
examples (also known as adversarial training)~\cite{goodfellow2014explaining,
shaham2015understanding, kurakin2016adversarial, madry2017towards,
tramer2017ensemble}. Adversarial training has proven successful in defending
against adversarial attacks on small images, especially on MNIST and
CIFAR-10~\cite{madry2017towards}. However, training on large-scale image
classification tasks remains an open challenge. On ImageNet,
Kurakin et al.~\cite{kurakin2016adversarial} report that adversarial training yields a robust classifier
against the FGSM attack, but adversarial training fails for the stronger iterative PGD attack.
Other efforts to robustify the model through retraining, including
modifying the network~\cite{ranjan2017improving},
regularization~\cite{cisse2017parseval} and data
augmentation~\cite{papernot2016distillation, zantedeschi2017efficient,
cisse2017parseval} have not shown to improve over adversarial training on ImageNet.

% Test-time only works well in blackbox
Another family of defenses do not require retraining. These
defenses pre-process the image with a transformation and then call a pre-trained
classifier to classify the image. For example, the JPEG
compression~\cite{dziugaite2016study}, image re-scaling~\cite{lu2017no,
xie2017mitigating}, feature squeezing~\cite{xu2017feature}, quilting~\cite{guo2017countering} and neural-based transformations~\cite{gu2014towards, meng2017magnet, liao2017defense, shen2017ape} are found to be effective in mitigating blackbox attacks.
Compared to retraining defenses, these defenses are computationally efficient and
easy to integrate with pre-trained classifiers. However, since they
are not crafted against whitebox attacks, their robustness rarely holds against adversaries
who have full knowledge of the transformation (see \cite{carlini2017magnet,athalye2018obfuscated} and Section~\ref{sec:imagenet} of this paper).

\begin{figure}
\centering
\begin{tabular}{ccc}
\includegraphics[height=0.13\textwidth]{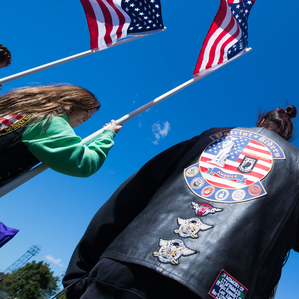}&
\includegraphics[height=0.135\textwidth]{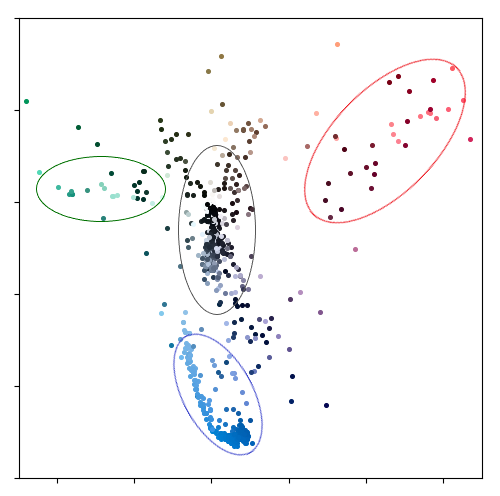}&
\includegraphics[height=0.13\textwidth]{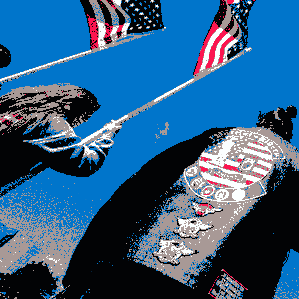}\\
(a) & (b) & (c)
\end{tabular}
\caption{(a) The original image. (b) Pixel distribution in the a-b axes of the Lab color space, which has a cluster structure. (c) Output of randomized discretization.}
\label{fig:flag_example}
\vspace{-10pt}
\end{figure}

In this paper, we present a defense that is specifically designed to work
against whitebox attacks without retraining.
%Our main contribution is to introduce a stochastic machinery to image transformation.
Our approach, \emph{randomized discretization} (or RandDisc), adds
Gaussian noise to each pixel, replaces each pixel with the closest cluster center,
and feeds the transformed image to any pre-trained classifier.
Figure~\ref{fig:flag_example} illustrates the intuition:
Image pixels in the color space often cluster. For pixels close
to a particular cluster center, their cluster assignments will be stable
under perturbations, and thus robust to the adversarial attack. For pixels
that are roughly equidistant from two centers,
their cluster assignments will be randomized by the injection of Gaussian noise,
which also mitigates the effect of the adversarial attack.

% Certificates
As argued in previous work \cite{raghunathan2018certified,kolter2017provable},
the gold standard for security is having a certificate that a defense provably works
against \emph{all} attacks.
In this work, we prove such a lower bound on the accuracy based on information-theoretic arguments:
Randomized discretization reduces the KL divergence between the
clean image and the adversarial image. If the two
distributions are close enough, then from an information-theoretic point of view,
no algorithm can distinguish the two transformed images. Thus,
the adversary cannot perturb the image to make significant modifications to the induced distribution over the classifier's
output. Previous defenses with robustness certificates~\cite{raghunathan2018certified, kolter2017provable}
require retraining and are feasible only on small-scale models.
In contrast, our defense requires no retraining and works on top any
pre-existing model.

% RandMix
Empirically, we evaluate a defense's whitebox robustness by performing the
iterative projected gradient descent (PGD) attack~\cite{madry2017towards}. If the
model is differentiable, then we generate adversarial examples by maximizing
its cross-entropy via PGD directly. Since RandDisc is non-differentiable,
we define a differentiable approximation of it called \emph{randomized
mixture} (or RandMix). We generate adversarial examples by attacking
RandMix, then use these examples to evaluate the robustness of both RandDisc
and RandMix.

% MNIST
Our experiments show that randomized discretization significantly improves a
classifier's robustness. On the MNIST dataset, our defense combined with a vanilla
CNN achieves 94.4\% accuracy on perturbations of $\ell_\infty$-norm
$\epsilon=0.1$, whereas the vanilla CNN achieves only 12.0\% accuracy without the
defense. In addition, the certified accuracy of the defense is consistently higher
than the empirical accuracy of the model without the defense.

% ImageNet
We also tested our approach on the subset of ImageNet used by the NIPS 2017 Adversarial Attacks \&
Defenses challenge~\cite{nips2017competition}. When integrated with a vanilla
InceptionResNet model, RandDisc achieves superior robustness compared to other
transformation-based
defenses~\cite{dziugaite2016study,xu2017feature,guo2017countering,xie2017mitigating}.
Its accuracy under the PGD attack is 3--5 times higher than the adversarially-trained
InceptionResNet model~\cite{tramer2017ensemble}.
Finally, we evaluate our defenses on the adversarial examples generated by the top 3
attacks according to the competition leaderboard. Compared to the top 3 winning
defenses, our defense obtains at least 18\% higher accuracy on average
and at least 35\% higher accuracy against the worst-case attack (for each defense).

\section{Background}

Let $x$ be an image with $n$ pixels. The $i$-th
pixel is represented by $x_i \in \R^q$,
where $q$ is the number of channels (1 for grayscale, 3 for color).
Let $x'$ denote the corresponding adversarially perturbed image,
which we assume to be $\epsilon$ close to $x'$ in $\ell_\infty$: $\norms{x-x'}_\infty \leq \epsilon$.
Earlier work has developed various of ways to defend against $\ell_\infty$-norm
attacks~\cite{goodfellow2014explaining, madry2017towards, tramer2017ensemble,
zantedeschi2017efficient, nips2017competition}.

Projected gradient descent (PGD)~\cite{madry2017towards} is an efficient algorithm to perform the
$\ell_\infty$-norm attack. Given a classifier and its loss
function $L$, The PGD computes iterative update:
\begin{align}\label{eqn:pgd}
	x_{t+1} \leftarrow \Pi\Big(x_t + \eta \cdot \mbox{sign}(\nabla_x L(x_t))\Big).
\end{align}
Here, $\eta$ is the step size, and $\Pi(\cdot)$ is the projection into the $\ell_\infty$-ball
$\{u:\norms{u-x}_\infty \leq \epsilon\}$. If we set $\eta = \epsilon$ and run PGD for only one iteration, then it is called the fast gradient sign method
(FGSM)~\cite{goodfellow2014explaining}.
Since PGD is a stronger attack than FGSM in the whitebox
setting~\cite{kurakin2016adversarial, kurakin2016world, carlini2017towards,
dong2017boosting}, we use PGD (that maximizes cross-entropy) to generate
adversarial examples.

We briefly mention other attacks: The Jacobian saliency map
attack~\cite{papernot2016limitations} corrupts the image by modifying a small
fraction of pixels. The DeepFool
attack~\cite{moosavi2016deepfool} computes the minimal perturbation necessary
for misclassification under the $L_2$-norm.
Carlini et al.~\cite{carlini2017towards} define an $\ell_\infty$ attack, which
can be viewed as running PGD to maximize a different loss, but its effect is
similar to maximizing the cross-entropy loss~\cite{madry2017towards, guo2017countering}.

%\myparagraph{Defenses} There is a line of work on the test-time
%defense~\cite{dziugaite2016study, lu2017no, xie2017mitigating, xu2017feature,
%guo2017countering, ehillon2017stochastic, buckman2018thermometer}. Although
%many of them propose discrete input transformations, and a few of them mention
%the importance of incorporating randomness~\cite{guo2017countering,
%ehillon2017stochastic}, none of them provides certificate against whitebox
%attacks. Empirically, the whitebox robustness of these methods are only
%evaluated on small images of MNIST and CIFAR-10. Our work provides a whitebox
%robustness certificate, and demonstrates its empirical robustness on ImageNet
%pictures.
%
%Two recent papers propose robust classifiers with provable
%guarantees~\cite{raghunathan2018certified, kolter2017provable}. These methods
%define an upper bound of the worst-case loss, or an outer bound of the set of
%adversarial examples. Then they optimize the model on these relaxations to
%obtain certificates. Due to the complexity of the relaxation techniques (which
%involve LP and SDP) and constraints on the network architecture, these methods
%are currently feasible only on small-scale classifiers on the MNIST dataset. In
%contrast, our defense is computationally efficient and can be integrated with
%arbitrarily complex classifiers.

\section{Stochastic transformations}
\label{sec:algorithm}

In this section, we present three defenses based on stochastic transformations of the image:
Gaussian randomization (Gaussian),
randomized discretization (RandDisc),
and randomized mixture (RandMix). The later two defenses  can be viewed as adding a (potentially stochastic) filter on top of the Gaussian randomization.
Each of these stochastic transformations produces an image $\xtilde$ with the
same size and the same semantics as $x$, so we can feed it directly into any \emph{base classifier}
and return the resulting label.

\myparagraph{Gaussian randomization}

The Gaussian randomization defense simply adds Gaussian
noise to raw pixels of the image. Given an image $x$, it constructs a noisy
image $\xtilde$ such that
\[
	\xtilde_i = x_i + w_i \quad \mbox{where } w_i\sim \sN(0, \sigma^2 I).
\]

\myparagraph{Randomized discretization}

The randomized discretization (RandDisc) defense first adds Gaussian noise to each
pixel and then discretizes each based on a finite number of cluster centers.
To define these clusters,
we randomly sample $s$ indices $(i_1,\dots,i_s) \in [n]^s$ with replacement.
Then we construct random vectors $\bvec \defeq (b_1,\dots,b_s)$ such that
\[
	b_j \defeq x_{i_j} + \varepsilon_j \quad \mbox{where } \varepsilon_j \sim \sN(0, \tau^2I).
\]
We then use any \emph{selection algorithm} to select $k$ cluster centers $\cvec \defeq
(c_1,\dots,c_k)$ from $\bvec$.
For example, it could draw $k$ points from $\bvec$ using the k-means++ initialization~\cite{arthur2007kmeans}, or
simply define $\cvec$ independent of $\bvec$.
%\pl{what about k-means? or give the algorithm we actually run; how does one choose?}
%\pl{motivate why two stage if we're just going to sample $k$ random points}

Once the cluster centers are chosen, we substitute each pixel by one of the centers:
\begin{align}\label{eqn:rand-discretization}
\xtilde_i = r(x_i|\cvec) \defeq \underset{c\in\{c_1,\dots,c_k\}}{\argmin} \ltwos{x_i + w_i - c},
\end{align}
 In other words, the stochastic function $r(x_i|\cvec)$ assigns the noisy pixel
 $x_i + w_i$ to its nearest neighbor.

\myparagraph{Randomized mixture} The randomized mixture (RandMix) defense
selects cluster centers in the same way as RandDisc, but rather than
choosing a single cluster center, it computes the weighted mean:
\begin{align}\label{eqn:rand-mixture}
\xtilde_i = m(x_i|\cvec) \defeq \frac{\sum_{j=1}^k c_j\cdot e^{-\alpha\ltwos{x_i+w_i-c_j}^2}}{\sum_{j=1}^k e^{-\alpha\ltwos{x_i+w_i-c_j}^2}},
\end{align}
$\alpha > 0$ is an inverse temperature parameter.
When $\alpha \to \infty$, RandMix converges to RandDisc.
Thus, RandMix can be viewed as an approximation to RandDisc.

Since the function $m(x_i|\cvec)$ is differentiable, we can use PGD to generate
adversarial examples for RandMix. Furthermore, since RandMix is an
approximation of RandDisc, we expect the attack to transfer to RandDisc, as
long as $\alpha$ is reasonably large. Athalye el al.~\cite{athalye2018obfuscated} show
that this approach successfully attacks established defenses.
We use it to evaluate RandDisc in Section~\ref{sec:experiment}.

\section{Theoretical analysis}
\label{sec:analysis}

Consider any two images $x$ and $x'$ satisfying $\norms{x-x'}_\infty \leq
\epsilon$.
Think of $x$ as a clean image and $x'$ as the adversarially perturbed image.
Let $\xtilde$ and $\xtilde'$ be the transformed versions of $x$ and
$x'$, generated by one of the transformations in Section~\ref{sec:algorithm}.
Suppose that $\xtilde$ is correctly classified
 by a base classifier $f$. In this section, we derive a
sufficient condition under which the pre-processed perturbed image, namely
$\xtilde'$ is also correctly classified by the same base classifier $f$.
This conclusion certifies the robustness of the defense.

For arbitrary random variable $\uvec$, we use $\prob_{\uvec}$ to denote its
probability distribution (or its density function if $\uvec$ is continuous). 
The KL divergence between two distributions $\prob_{\uvec}$
and $\prob_{\uvec'}$ is defined as:
\[
	\KL{\prob_{\uvec}\|\prob_{\uvec'}} \defeq \int \prob_{\uvec}(u) \log\Big(\frac{\prob_{\uvec}(u)}{\prob_{\uvec'}(u)}\Big) du.
\]
We also define the total variation distance between two distributions:
\[
	\tv{\prob_{\uvec} - \prob_{\uvec'}} \defeq \frac{1}{2} \int |\prob_{\uvec}(u) - \prob_{\uvec'}(u)| du.
\]

\subsection{Upper bound on the KL divergence}
\label{sec:kl-divergence}

A critical part of our analysis is an upper bound on the KL divergence between
the distributions of $\xtilde$ and $\xtilde'$. We derive this bound for the
Gaussian randomization defense and the RandDisc defense.

\myparagraph{Gaussian randomization} Using the decompositionality of KL
divergence and the formula of KL divergence between two normal distributions, we have:
\begin{align}\label{eqn:kl-bound-gaussian}
\KL{\prob_{\xtilde}\|\prob_{\xtilde'}} &= \sum_{i=1}^n \KL{\prob_{\xtilde_i}\|\prob_{\xtilde'_i}} = \sum_{i=1}^n \frac{\ltwos{x_i - x'_i}^2}{2\sigma^2}\nonumber\\
& \leq \frac{n q \epsilon^2}{2\sigma^2},
\end{align}
where the last inequality uses the fact that $\norms{x_i - x'_i}_\infty \leq \epsilon$.

\myparagraph{RandDisc} RandDisc adds a discrete filter on top of the
Gaussian randomization, which further reduces the KL divergence. The following
proposition presents an upper bound. The proposition is proved by using the
properties of KL divergence and a \emph{data processing inequality}.

\begin{proposition}\label{prop:kl-upper-bound}
Given hyperparameters $(s,k,\sigma,\tau)$ and the number of channels $q$. the
KL divergence between $\xtilde$ and $\xtilde'$ can be upper bounded by
\begin{align}\label{eqn:kl-combined}
&\KL{\prob_{\xtilde}\|\prob_{\xtilde'}} \leq \frac{s q \epsilon^2}{2\tau^2} + \sum_{i=1}^n  \sup_{\Delta \in [-\epsilon, \epsilon]^q}\nonumber \\
&\qquad\qquad\E_{u\sim \prob_{\cvec}}\Big[ 
\KL{\prob_{r(x_i|u)}\|\prob_{r(x_i+\Delta|u)}} \Big].
\end{align}
\end{proposition}

By Proposition~\ref{prop:kl-upper-bound}, we compute the upper bound
by sampling ``cluster centers'' $u$, then compute the KL divergence between two multinomial distributions (the
distributions of $r(x_i|u)$ and $r(x_i+\Delta|u)$). 
If the multinomial distributions have no closed form, we draw $r(x_i|u)$ and $r(x_i+\Delta|u)$ many times, then compute the KL divergence between the two empirical distributions (see \cite{han2016minimax} for a more efficient estimator for the KL divergence). 

The remaining problem is to compute the supremum over $\Delta\in [-\epsilon,\epsilon]^q$. We note that the supremum is
defined on a smooth function of $\Delta$. As a result, the function can be approximated by an expansion
$g^\top \Delta + O(\Delta^2)$ for some gradient vector $g$. If $\epsilon$ is small enough (which is the case for our
setting), then the supremum must be achieved at a point where $g^\top \Delta$ is maximized, and therefore must 
be at the vertices of $[-\epsilon,\epsilon]^q$. If $q = 3$, it can be computed by
simply enumerating the $2^3 = 8$ possibilities.

If the relation $\E_{u\sim \prob_{\cvec}}[\KL{\prob_{r(x_i|u)}\|\prob_{r(x_i+\Delta|u)}}] \ll \frac{q\epsilon^2}{2\sigma^2}$ holds on many pixels, then by comparing the right-hand side of \eqref{eqn:kl-bound-gaussian} and \eqref{eqn:kl-combined}, we find that RandDisc's upper bound will be much smaller than that of the Gaussian randomization. This happens if both $x_i$ and $x_i+\Delta$ are close to a particular cluster center, so that the distributions of both $r(x_i|u)$ and $r(x_i+\Delta|u)$ concentrate on the same point. We found that this property holds on MNIST and ImageNet images.

\subsubsection{Proof of Proposition~\ref{prop:kl-upper-bound}}\label{sec:proof-kl-upper-bound}

\paragraph{Notations.}

Let $\prob_{\vvec|{\bf u}}(\cdot|u)$ be the distribution of an arbitrary random variable $\vvec$ conditioning on
the the event ${\bf u} = u$ of random variable $\bf u$. We define the KL divergence between two conditional distributions $\prob_{\vvec|\uvec}$ and $\prob_{\vvec'|\uvec'}$ as:
\begin{align*}
\KL{\prob_{\vvec|\uvec}\|\prob_{\vvec'|\uvec'}} \defeq \E_{u\sim p_\uvec}\Big[\KL{\prob_{\vvec|\uvec}(\cdot|u)\|\prob_{\vvec'|\uvec'}(\cdot|u)}\Big].
\end{align*}
The \emph{chain rule} of KL divergence states that:
\[
	\KL{\prob_{\uvec,\vvec}\|\prob_{\uvec',\vvec'}} = \KL{\prob_{\uvec}\|\prob_{\uvec'}} + \KL{\prob_{\vvec|\uvec}\|\prob_{\vvec'|\uvec'}}.
\]
It implies the following inequality:
\begin{align}
	 \KL{\prob_{\vvec}\|\prob_{\vvec'}} &\leq \KL{\prob_{\vvec}\|\prob_{\vvec'}} + \KL{\prob_{\uvec|\vvec}\|\prob_{\uvec'|\vvec'}}\nonumber\\
	& = \KL{\prob_{\uvec,\vvec}\|\prob_{\uvec',\vvec'}}\nonumber\\
& =  \KL{\prob_{\uvec}\|\prob_{\uvec'}} + \KL{\prob_{\vvec|\uvec}\|\prob_{\vvec'|\uvec'}}.
	\label{eqn:kl-chain}
\end{align}
As a special case, if $\uvec\to\vvec$ and $\uvec'\to\vvec'$ are processed by the same channel, then $\KL{\prob_{\vvec|\uvec}\|\prob_{\vvec'|\uvec'}}=0$, so that we have the following \emph{data processing inequality}:
\[
	\KL{\prob_{\vvec}\|\prob_{\vvec'}} \leq \KL{\prob_{\uvec}\|\prob_{\uvec'}}.
\]

\paragraph{Proof}

We use $(\ivec,\bvec, \cvec)$ and $(\ivec',\bvec',
\cvec')$ to represent the vectors formed in the process of transforming $x$ to $\xtilde$ and
transforming $x'$ to $\xtilde'$ (see Section~\ref{sec:algorithm}). By inequality~\eqref{eqn:kl-chain}:
\begin{align*}
	 \KL{\prob_{\xtilde}\|\prob_{\xtilde'}} \leq \KL{\prob_{\cvec}\|\prob_{\cvec'}} + \KL{\prob_{\xtilde|\cvec}\|\prob_{\xtilde'|\cvec'}}.
\end{align*}
Since the vectors $\cvec$ and $\cvec'$ are selected by the same algorithm given $\bvec$ and $\bvec'$, the data processing inequality implies:
\[
	\KL{\prob_{\cvec}\|\prob_{\cvec'}} \leq \KL{\prob_{\bvec}\|\prob_{\bvec'}} = \sum_{j=1}^s \KL{\prob_{b_j}\|\prob_{b'_j}},
\]
where the last equation uses the fact that the sub-vectors of $\bvec$ and $\bvec'$ are independent.
Note that the sub-pixels of $\xtilde$ (conditioning on $\cvec$) and $\xtilde'$ (conditioning on $\cvec'$) are also independent, thus we have:
\begin{align}\label{eqn:kl-decomposition}
&\KL{\prob_{\xtilde}\|\prob_{\xtilde'}} \nonumber\\
&\qquad \leq \sum_{j=1}^s \KL{\prob_{b_j}\|\prob_{b'_j}} + \sum_{i=1}^n \KL{\prob_{\xtilde_i|\cvec}\|\prob_{\xtilde'_i|\cvec'}}.
\end{align}
For the first term on the right-hand size, inequality~\eqref{eqn:kl-chain} implies:
\[
\KL{\prob_{b_j}\|\prob_{b'_j}} \leq \KL{\prob_{i_j}\|\prob_{i'_j}} + \KL{\prob_{b_j|i_j}\|\prob_{b'_j|i'_j}}.
\] 
Since $\KL{\prob_{i_j}\|\prob_{i'_j}}$ is zero and $\KL{\prob_{b_j|i_j}\|\prob_{b'_j|i'_j}}$ is 
bounded by $\frac{q \epsilon^2}{2\tau^2}$, we have:
\begin{align}\label{eqn:kl-for-c}
\KL{\prob_{b_j}\|\prob_{b'_j}} \leq \frac{q \epsilon^2}{2\tau^2}.
\end{align}
For the second term on the right-hand side of~\eqref{eqn:kl-decomposition}, using the fact that $x_i$ is $\epsilon$-close to $x'_i$ in the $\ell_\infty$-norm, we have:
\begin{align}\label{eqn:kl-for-pixel}
&\KL{\prob_{\xtilde_i|\cvec}\|\prob_{\xtilde'_i|\cvec'}} \nonumber \\
&\qquad \leq \sup_{\Delta\in [-\epsilon, \epsilon]^q} \E_{u\sim \prob_\cvec}[
\KL{\prob_{r(x_i|u)}\|\prob_{r(x_i+\Delta|u)}}],
\end{align}
where the function $r$ is defined in Section~\ref{sec:algorithm}. Combining inequalities~\eqref{eqn:kl-decomposition}--\eqref{eqn:kl-for-pixel}, we obtain:
\begin{align*}
&\KL{\prob_{\xtilde}\|\prob_{\xtilde'}} \leq \frac{s q \epsilon^2}{2\tau^2} + \sum_{i=1}^n  \sup_{\Delta \in [-\epsilon, \epsilon]^q} \\
&\qquad \E_{u\sim \prob_{\cvec}}\Big[ \Big( 
\KL{\prob_{r(x_i|u)}\|\prob_{r(x_i+\Delta|u)}} \Big)\Big],
\end{align*}
which completes the proof.

\subsection{Certified accuracy}
\label{sec:certificate}

Next, we translate the KL divergence bound (between transformed image distributions) to a bound on classification accuracy.  Consider an arbitrary base
classifier $f: \R^n \to \N$ and let $y\in \N$ be the ground truth label. We
define the \emph{margin of classification} to be the probability of predicting
the correct label $y$, subtracted by the maximal probability of predicting a
wrong label. Formally,
\begin{align}\label{eqn:margin-definition}
\mbox{margin}(\xtilde,y,f) \defeq P(f(\xtilde)=y) - \max_{z\neq y}P(f(\xtilde)=z).
\end{align}
Let $\xtilde$ and $\xtilde'$ be two images transformed from $x$ and $x'$. By
simple algebra, the margin can be related to the total variation distance:
\[
|\mbox{margin}(\xtilde',y,f) - \mbox{margin}(\xtilde,y,f)| \leq 2\, \tv{\prob_{f(\xtilde)} - \prob_{f(\xtilde')}}.
\] 
Using Pinsker's inequality and the data processing inequality (see Section~\ref{sec:proof-kl-upper-bound}), we obtain:
\begin{align*}
&|\mbox{margin}(\xtilde',y,f) - \mbox{margin}(\xtilde,y,f)|\\
&\qquad \leq \sqrt{2\, \KL{\prob_{f(\xtilde)}\|\prob_{f(\xtilde')}}} \\
&\qquad \leq
	\sqrt{2\, \KL{\prob_{\xtilde}\|\prob_{\xtilde'}}}.
\end{align*}
Equivalently, the above inequality implies:
\begin{align*}
\mbox{margin}(\xtilde',y,f) \geq \mbox{margin}(\xtilde,y,f) - \sqrt{2\, \KL{\prob_{\xtilde}\|\prob_{\xtilde'}}}.
\end{align*}
This implies that if $\KL{\prob_{\xtilde}\|\prob_{\xtilde'}}$ is strictly smaller
than $\frac{1}{2}(\mbox{margin}(\xtilde,y,f))^2$, then the classifier's probability of making
the correct prediction will be greater than any incorrect probability. Consequently, we can evaluate
$f(\xtilde')$ multiple times and use majority voting to retrieve the correct
label. By applying the union bound and Hoeffding's inequality,
we obtain the following proposition:

\begin{proposition}\label{theorem:certified-correctness}
Consider running a Gaussian defense or a RandDisc defense 
on a $k$-category classification instance for $m$ independently times.
Let $U_{\rm
  KL}(x)$ be the KL divergence upper bound obtained
  from~\eqref{eqn:kl-bound-gaussian} or~\eqref{eqn:kl-combined}. If 
  \begin{align}\label{eqn:certify_condition}
 \delta \defeq U_{\rm KL}(x) - \frac{1}{2}(\mbox{margin}(\xtilde,y,f))^2 > 0,
 \end{align}
then the most frequent output label is correct with probability at least $1-ke^{-2m\delta^2}$.
\end{proposition}

We define the \emph{certified accuracy} of a defense to be the fraction of
examples that satisfy the condition~\eqref{eqn:certify_condition}.
It can be numerically computed by computing the KL divergence bound \eqref{eqn:kl-bound-gaussian} or~\eqref{eqn:kl-combined}, and the margin of classification \eqref{eqn:margin-definition}.

\section{Experiments}
\label{sec:experiment}

In this section, we evaluate our defenses against whitebox projected gradient descent (PGD) attacks on the MNIST and the ImageNet datasets. We follow Athalye et al.~\cite{athalye2018obfuscated} to construct the strongest possible attack. In particular, if the defense is non-differentiable, then we attack a differentiable approximation of it. If the defense is stochastic, then in each round we average multiple independent copies of the gradients to perform the PGD. The combination of these techniques was shown to successfully attack many existing defenses.

For the cluster selection algorithm of RandDisc and RandMix, we use an adaptive sampling algorithm that mimics the k-means++ initialization but doesn't involve the non-differentiable 
updates of k-means\footnote{For evaluation purpose, we make the defense differentiable, so that it can be effectively attacked by PGD. In practice, one could use the iterative k-means++ to obtain better clusters.}. Each cluster center $c_j$ is sampled from $\{b_1,\dots,b_s\}$ under the following distribution:
\[
	P(c_j = b) \propto e^{\gamma \min_{\ell\in\{1,\dots,j-1\}} \ltwos{c_{\ell} - b}^2},~~b\in \{b_1,\dots,b_s\}.
\]
By construction, points that are far from the existing centers
$\{c_1,\dots,c_{j-1}\}$ will be more likely to be selected. Assuming that the pixel values belong to $[0,t]$, we set $s=100$ and $\gamma=\frac{40}{t^2}$ for the selection algorithm, and set $\alpha = \frac{40}{t^2}$ for RandMix to make sure that it is differentiable.

\subsection{MNIST}

The MNIST test set consists of 10,000 examples. We use the normally-trained CNN of \cite{madry2017towards} as the base
classifier. The model reaches 99.2\%
accuracy on the test set. Each pixel in the input image is normalized to interval $[0,1]$. All our
defenses select $k=2$ clusters (because the images are black-and-white) and use $\tau=\sigma=0.15$ as the noise scale. The adversary runs 40 iterations of PGD. We compute 20 independent
copies of the stochastic gradient in each iteration of PGD, and take their mean
to execute one PGD step. We attack RandMix as a proxy to evaluate the
robustness of RandDisc.

\myparagraph{Results}

\begin{figure}
\centering
\includegraphics[width=0.35\textwidth]{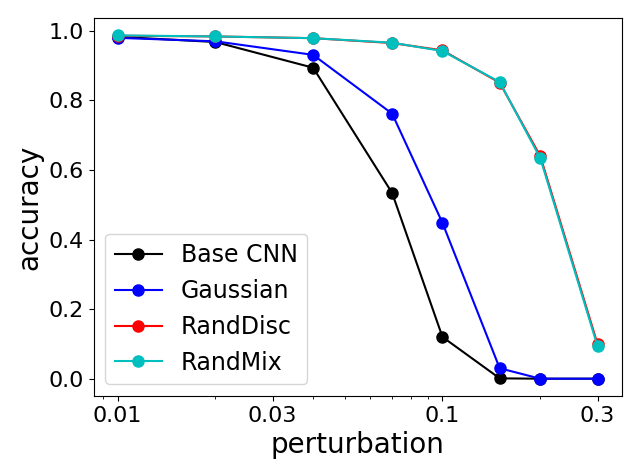}
\caption{Accuracies under a whitebox attack on MNIST.
  The plot compares the base CNN (no defense), the Gaussian defense, the RandDisc defense and the RandMix defense.}
\label{fig:mnist_empirical_study}
\end{figure}

The results under whitebox attack are presented in
Figure~\ref{fig:mnist_empirical_study}. The Gaussian defense consistently
outperforms the base classifier, confirming the benefits of randomization. The
difference is significant on $\epsilon\in\{0.07, 0.1\}$. Both RandDisc and
RandMix (whose curves overlap) are substantially more accurate than Gaussian. In particular, RandDisc
achieves 94.4\% accuracy for $\epsilon=0.1$, which
exceeds the 84\% accuracy of \cite{raghunathan2018certified}, the 91\%
accuracy of \cite{raghunathan2018semidefinite}
and the 93.8\% accuracy of \cite{kolter2017provable}. 
These later models provide theoretical certificates, but require re-training the model.
We also note that Gowal et al.~\cite{gowal2018effectiveness} proposed 
an interval bound propagation method to derive a tighter upper bound
on the worst-case loss. Optimizing this relaxation achieves 97.7\% certified
accuracy on MNIST. 

Interestingly, on the adversarially-trained (against PGD)
model of \cite{madry2017towards}, our stochastic defenses actually
hurt robustness. This might be due to the fact that the adversarially-trained model
is crafted for the original input distribution. The model is already quite robust, 
achieving 92.7\% accuracy under the PGD attack for $\epsilon=0.3$. 
Our defense modifies the input distribution, and consequently slightly lowers this
particular model's accuracy which is optimized for the original 
distribution.

\myparagraph{Certified accuracy}

\begin{table}
\centering
\small
\begin{tabular}{|l|c|c|}
\hline
condition & filter size & stride \\\hline
$\epsilon \in (0,0.02]$ & 1 & 1 \\\hline
$\epsilon \in (0.02,0.05]$ & 1 & 2 \\\hline
$\epsilon \in (0.05,0.07]$ & 2 & 3 \\\hline
$\epsilon \in (0.07,0.1]$ & 2 & 4 \\\hline
$\epsilon \in (0.1,0.3]$ & 2 & 7 \\\hline
\end{tabular}
\vspace{10pt}
  \caption{For each adversary condition $\epsilon$, this table
  shows the filter size and stride of the downsampling operator.}
\label{table:max-pooling-params}
\end{table}

\begin{figure}
\centering
\small
\begin{tabular}{cc}
\includegraphics[width=0.2\textwidth]{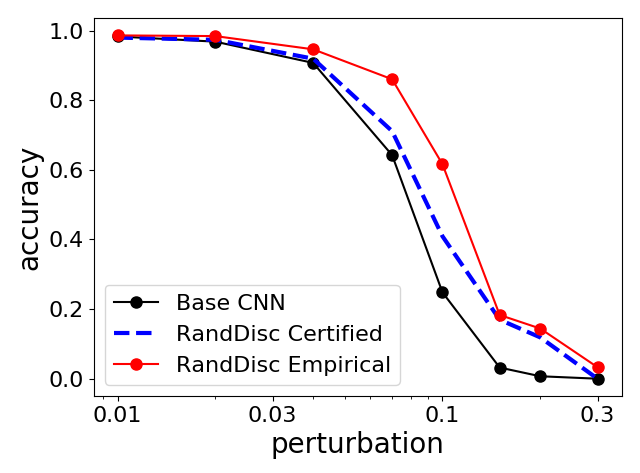}& 
\includegraphics[width=0.2\textwidth]{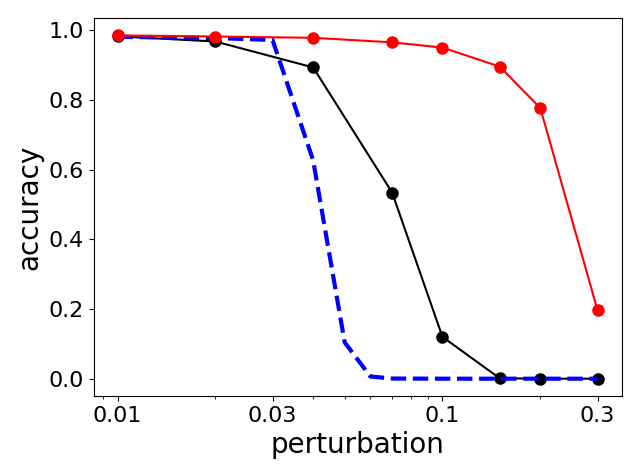}\\
(a) with downsampling & (b) without downsampling
\end{tabular}
\caption{Certified and empirical accuracy of RandDisc under the whitebox attack on MNIST.}
\label{fig:mnist_certified_accuracy}
\end{figure}

The Gaussian randomization defense, due to its overly
loose KL divergence bound, has very low certified accuracy compared to its empirical accuracy against
white-box attacks.
See Section~\ref{sec:kl-divergence} for an explanation of why its KL divergence bound 
can be much worse than that of RandDisc. Thus we focus on the certified accuracy of RandDisc.

We report results on the vanilla RandDisc as well as a down-sampled variant of it.
For the down-sampled version, given image $x$, we feed it into a max-pooling layer to obtain a smaller image $x_s$. 
Then we feed $x_s$ to RandDisc, then resize it back to 28x28 using bilinear interpolation before feeding to the base classifier. Note that max-pooling never increases the perturbation scale. Thus, the adversary can perturb $x_s$ by at most 
 $\epsilon$. 
 
To analyze this variant, we treat $x_s$ as the input image,
and consider the following new base classifier: it is the pipeline 
consisting of bilinear interpolation and the old base classifier. The new base
classifier takes down-sampled images $x_s$ as input.
As a consequence, we can use the same technique to compute the certified
accuracy of this defense. The only difference is that the 
KL divergence bound is computed on the smaller
image $x_s$ instead of $x$. Since $x_s$ has fewer pixels, it results in a lower KL divergence
bound, and thus a higher certified accuracy provided the down-sampling parameters are properly chosen. The filter
size and the stride are selected as a function of the perturbation scale, as reported in
Table~\ref{table:max-pooling-params}.

Figure~\ref{fig:mnist_certified_accuracy} plots the certified accuracy under various perturbation scales.
With down-sampling, the certified accuracy is consistently higher than the
accuracy of the base classifier, confirming that the worst-case performance of the defense
is better than the empirical performance of the model without the defense.
Without down-sampling, the certified accuracy decreases, but the empirical accuracy
increases. This suggests that the theoretical lower bound still has much room for improvement.

\subsection{ImageNet}
\label{sec:imagenet}

We report experiments on a subset of ImageNet used
by the NIPS 2017 Adversarial Attacks \& Defenses challenge~\cite{nips2017competition}. The dataset contains 1,000
images from 1,001 categories.
Each RGB channel is an integer between 0 and 255. 
We use a pre-trained
\emph{InceptionResNet-V2} model~\cite{szegedy2017inception} as the base
classifier. The classifier is trained on the standard ImageNet training set. We
note that the competition has removed images 
that were frequently misclassified by state-of-the-art neural models even
without perturbation. The base classifier achieves 99.3\% accuracy 
on the resulting dataset.

We compare our defenses with four transformation-based defenses that require no retraining:
\begin{itemize}[leftmargin=15pt]
\item \textbf{BitDepth}~\cite{xu2017feature}: reduce the bit-depth of each RGB channel from 8 bits to 2 bits by quantification.
\item \textbf{JPEG}~\cite{dziugaite2016study}: JPEG compression and decompression with a compression ratio of $4$.
\item \textbf{Total variation minimization (TVM)}~\cite{guo2017countering}: compute a denoised image $z$ by minimizing the following objective function:
\[
	\ell(z) \defeq \frac{1}{2}\ltwos{z-x}^2 + \lambda\cdot{\rm TV}(z),
\]
where ${\rm TV}(z)$ stands for the total variation of image $z$. We choose
    $\lambda=0.2$ and perform 20 iterations of gradient descent to approximate
    the minimization.

\item \textbf{ResizePadding}~\cite{xie2017mitigating}: resize the image to a
  random size between 310 and 331, then pads zeros around the image in a random
    manner, so that the final image is of size $331\times 331$. 
\end{itemize}

The adversary generates adversarial examples by running 10 iterations of PGD
on each image to attack the respective defense. To attack BitDepth, JPEG and TVM, we define their differentiable approximations by replacing step functions by $\sigma(\alpha t)$, where $\sigma$ is the logistic function and $\alpha$ is the same as in RandMix. 

The Gaussian defense uses noise level $\sigma=32$. Both RandDisc and RandMix select $k=5$
cluster centers per image, and use $\tau=\sigma=32$. For attacking stochastic
models, we compute 100 independent copies of the stochastic gradients in each
iteration of PGD, then use their empirical mean to execute the PGD step. This
results in a strong whitebox attack at the cost of 100x computation time.

\begin{table*}
\centering
\small
\begin{tabular}{cc}
\begin{tabular}{|l|c|c|c|}
\hline
& $\epsilon=1$ & $\epsilon=2$ & $\epsilon=4$\\\hline
Base Classifier & 16.0\% & 6.0\% & 2.3\% \\\hline
BitDepth & 20.4\% & 8.7\% & 3.4\% \\\hline
JPEG & 22.4\%	& 9.9\%	& 4.5\% \\\hline
TVM & 25.5\% & 10.2\% & 4.2\% \\\hline
ResizePadding  & 30.7\%	& 9.8\%&	2.3\%\\\hline\hline
Gaussian  & 57.3\% & 29.9\% & 11.4\% \\\hline
RandDisc  & 56.0\% & 46.0\% & {\bf 29.4\%} \\\hline
RandMix  & {\bf 60.7\%} & {\bf 48.3\%} & 28.3\% \\\hline
\end{tabular}
&
\begin{tabular}{|l|c|c|c|c|}
\hline
Attack: & 1st & 2nd & 3rd & Mean \\\hline 
1st defense & {\bf 87.3\%} & 44.1\% & 33.8\% & 55.1\%\\\hline
2nd defense & 57.3\% & 28.3\% & 63.9\% & 49.8\%\\\hline
3rd defense & 54.7\% & 27.7\% & 61.1\% & 47.8\%\\\hline
BitDepth & 61.2\% & 61.3\% & 69.9\% & 64.0\%\\\hline
JPEG & 65.0\% & 51.0\% & 21.3\% & 45.8\% \\\hline
TVM & 58.6\% & 52.0\% & 62.6\% & 57.7\% \\\hline\hline
Gaussian  & 69.1\% & 66.3\% & 69.1\% & 68.2\% \\\hline
RandDisc  & 72.4\% & {\bf 69.0\%} & 75.9\% & {\bf 72.4\%} \\\hline
RandMix  & 69.8\% & 67.2\% & {\bf 76.2\%} & 71.1\% \\\hline
\end{tabular}\\\\
  (a) Against whitebox PGD attack & (b) Against winning attacks of NIPS 2017 competition
\end{tabular}
\vspace{10pt}
  \caption{Accuracies on the ImageNet subset of 1,000 images (NIPS Adversarial Attacks \& Defenses competition).}
\label{table:imagenet-tables}
\end{table*}

\myparagraph{Results} We evaluate all defenses on perburbation scales $\epsilon\in \{1,2,4\}$. The results are
reported on Table~\ref{table:imagenet-tables}(a). Both
RandDisc and RandMix consistently outperform other defenses, demonstrating the
benefit of combining randomization and discretization. 
For the rest of this subsection, we study the impact of various factors on
accuracy. For each setting of hyperparameters (e.g., noise scale and cluster number), we generate adversarial examples with respect
to those hyperparameters.
When the results of both RandDisc and
RandMix are available, we report the result of RandMix.

\begin{figure*}
\centering
\small
\begin{tabular}{cccc}
\hspace{-20pt}
\includegraphics[width=0.23\textwidth]{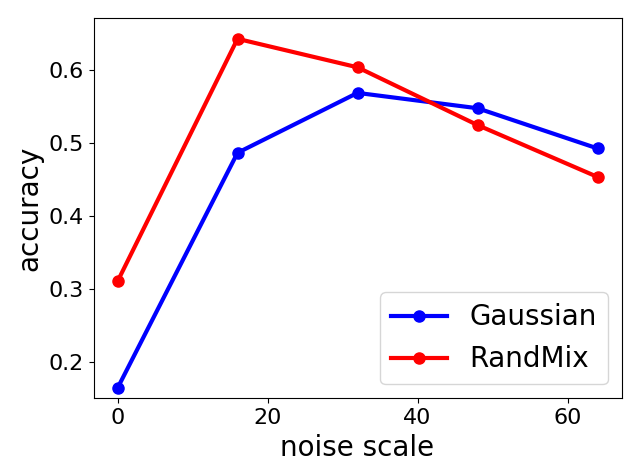}&
\includegraphics[width=0.23\textwidth]{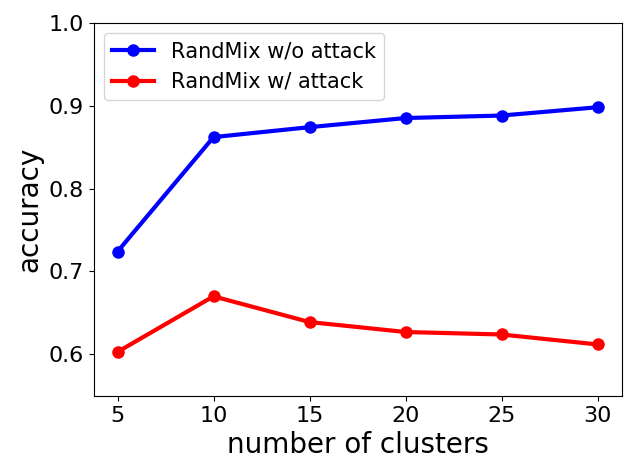}&
\includegraphics[width=0.23\textwidth]{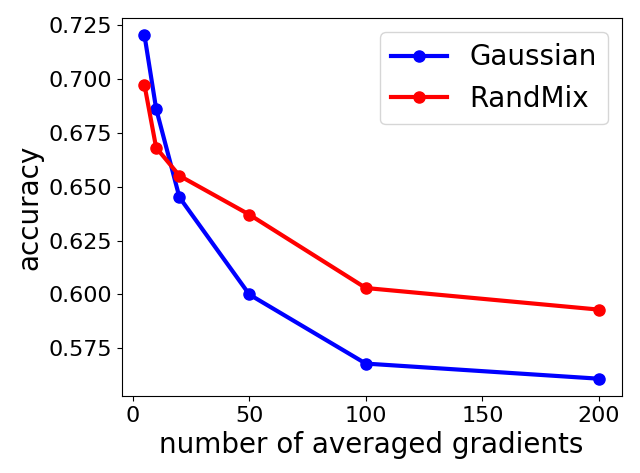}&
\includegraphics[width=0.23\textwidth]{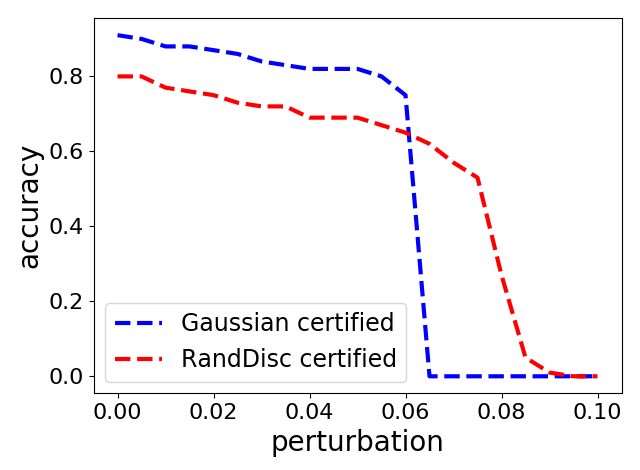}\\
(a) effect of noise & (b) effect of clusters & (c) effect of stochastic grad. & (d) certified accuracy
\end{tabular}
\caption{
  On the ImageNet subset (perturbation
  scale set to $\epsilon=1$ for (a)--(c)).
  (a), (b): adding noise or using less clusters increases robustness, but hurts accuracy on clean images; (c): the marginal utility of averaging more stochastic gradients diminishes; (d): Gaussian randomization and RandDisc are provably robust for small enough $\epsilon$.}
\label{fig:imagenet-plots}
\end{figure*}

\myparagraph{Effect of noise} To understand the effect of noise, we very the
noise parameter $\sigma$ from 0 to 64. The accuracies are plotted in
Figure~\ref{fig:imagenet-plots}(a). We find that both methods hit very low 
accuracies when there is no noise ($\sigma=0$). This confirms that the random noise is important for our defense. 
The maximal accuracy of RandMix is 7.4\% higher than that
of Gaussian randomization, meaning that the discretization filter helps
even if the noise scale is optimized.

\myparagraph{Effect of clusters} In Figure~\ref{fig:imagenet-plots}(b), we plot
the accuracy as a function of the number of clusters $k$. With more
clusters, the classifier's accuracy on clean images increases, but its
robustness drops. There is a certain point where the accuracy and the
robustness reach an optimal trade-off.

An alternative cluster selection algorithm is to make cluster
centers image-independent. To understand this approach, we select eight fixed colors of $\{64,192\}^3$ as pre-defined RGB
centers to evaluate RandMix. With these pre-defined cluster centers, the
accuracy of RandMix drops from (60.3\%, 48.8\%, 30.5\%) to (51.7\%, 33.6\%, 15.9\%) for
$\epsilon\in\{1,2,4\}$, respectively. This highlights the necessity of our adaptive clustering algorithm.

\myparagraph{Effect of stochastic gradients} Figure~\ref{fig:imagenet-plots}(c) shows that 
averaging more gradients indeed make the 
attack much stronger, but on both Gaussian and RandMix, 
the marginal utility diminishes. % as the number of gradients goes to infinity.
The decision of averaging 100 gradients is made based on a trade-off between
optimizing the attacking strength given our computational
resources.

\myparagraph{Effect of loss function} Our attack maximizes the cross-entropy loss.
An alternative loss function is proposed by Carlini and Wagner~\cite{carlini2017towards}, which target at mis-classification with high confidence. On this alternative loss, we rerun
the PGD attack with confidence parameter $\kappa=50$ (same as \cite{madry2017towards}). RandMix achieves accuracies 61.0\%, 49.0\%
and 26.4\% on $\epsilon=\{1,2,4\}$, respectively. This is similar to the numbers in Table~\ref{table:imagenet-tables}(a), confirming that our defense is stable under attacks optimizing different losses.

\myparagraph{Adversarially-trained model} If we take an adversarially trained
model as the base classifier, then our defense can be even stronger.
Specifically, we take an InceptionResNet-V2 model adversarially-trained against
the FGSM attack\footnote{\cite{tramer2017ensemble} report that the adversarial training 
fails against the PGD attack.}~\cite{tramer2017ensemble}. The model itself is vulnerable to
the iterative PGD attack: its whitebox accuracies are (18.4\%, 10.7\%, 5.8\%)
on $\epsilon\in\{1,2,4\}$. However, after integrating with RandMix, the
accuracies increase dramatically to (62.9\%, 54.2\%, 39.5\%), surpassing the
best accuracies (Table~\ref{table:imagenet-tables}(a)). This observation is the opposite
of that on MNIST, where our stochastic defenses actually hurt the adversarially-trained model.
We speculate that the adversarially-trained model on ImageNet is much weaker compared to the one on MNIST,
and thus our stochastic defenses have room to improve.

\myparagraph{Certified accuracy} We compute the certified accuracies
of Gaussian randomization and RandDisc, following the method of
Section~\ref{sec:analysis}. The KL divergence bound for RandDisc is about 1/3 of
that for the Gaussian randomization, 
but as Figure~\ref{fig:imagenet-plots}(d) shows, the
certified accuracy of both defenses are non-vacuous only on very small perturbations
($\epsilon < 0.1$). This is due to the fact that our KL divergence bound is the sum of the
KL divergence bounds on individual pixels. Since ImageNet is much higher
resolution than MNIST, it leads to a loose cumulative bound.

\myparagraph{Results on the NIPS 2017 competition}

Finally, we evaluated our defenses against the strongest attacks in
the NIPS 2017 Adversarial Attacks \& Defenses
competition~\cite{nips2017competition}. We downloaded the source code of the top 3 attacks on the final
leaderboard to generate adversarial examples. The perturbation scale is set to
$\epsilon=8$. On these examples, we compare our defenses with the top 3
defenses on the final leaderboard.

As the base classifier,
we use the adversarially-trained InceptionResNet-V2 model
obtained through ensemble adversarial training against
the FGSM attack, which is publicly available~\cite{tramer2017ensemble}. Since
the NIPS attacks are generally weaker than the whitebox attack, we use a lower
noise level $\tau=\sigma=16$, and $k=10$ clusters to preserve more information about
the image.
We also report BitDepth, JPEG and TVM, but ignore ResizePadding because the 2nd
defense is exactly the same as ResizePadding.

The results are reported in Table~\ref{table:imagenet-tables}(b). We notice that
the first defense is particularly strong against the first attack. It is potentially
due to the fact that they are submitted by the same team, and according to the
team's report~\cite{liao2017defense}, the defense is trained explicitly to
eliminate that particular attack. Nevertheless, RandDisc is at least 35\%
better than the top 3 defenses in the worst case, and at least 18\% better in
the average case.

On clean images, RandDisc and RandMix exhibit lower accuracies (88.6\% and 92.7\%) than the base classifier (97.1\%). This is due to the effect of the random noise and the discretization.
An open challenge is to make RandDisc robust against adversarial attacks without sacrificing the base classifier's clean-image accuracy.

\section{Discussion}

The idea of injecting randomness has been explored by previous work on robust classification. Gu et al.~\cite{gu2014towards} study injecting Gaussian noise as a blackbox defense for MNIST, but they report disappointing results.
This is consistent with our observation on MNIST that the Gaussian randomization's robustness is significantly weaker than that of RandDisc. Guo et al.~\cite{guo2017countering} propose that certain stochastic transformations can help defend against greybox attacks. Dhillon et al.~\cite{ehillon2017stochastic} propose to use stochastic activation functions, and evaluate the model on CIFAR-10. However, none of them demonstrates whitebox robustness on ImageNet.
Buckman et al.~\cite{buckman2018thermometer} propose a discrete transformation based on one-hot encodings, and report its whitebox robustness on MNIST and CIFAR-10. However,
Athalye et al.~\cite{athalye2018obfuscated} show that it can be broken by attacking the differentiable approximation. Kannan~\cite{kannan2018adversarial} proposed the logit pairing defense. They show that logit pairing exhibits better robustness than
adversarial training on ImageNet. These results are based on defending against targeted attacks, while in our experiment setting,
we defend against untargeted attacks.

In this paper, we have proposed randomized discretization as a defense, and have tried our best to attack it to empirically evaluate its robustness on ImageNet.
%We will release the code and welcome the community to launch stronger attacks against it. 
Our theoretical analysis provides an information-theoretic perspective to understanding stochastic defenses, and is complementary to existing certificates for deterministic classifiers, which rely on optimizing a relaxation of the worst-case loss~\cite{raghunathan2018certified,kolter2017provable,raghunathan2018semidefinite,gowal2018effectiveness}.

\paragraph{Reproducibility.} Code, data, and experiments
for this paper are available on the CodaLab platform:
\url{https://worksheets.codalab.org/worksheets/0x822ba2f9005f49f08755a84443c76456/}.

\bibliography{bib}
\bibliographystyle{abbrv}

\end{document}